\documentclass[conference]{IEEEtran}
\IEEEoverridecommandlockouts
\usepackage{cite}
\usepackage{amsmath,amssymb,amsfonts}
\usepackage{algorithmic}
\usepackage{graphicx}
\usepackage{textcomp}
\usepackage{xcolor}
\usepackage{float}
\usepackage{hyperref}

\def\BibTeX{{\rm B\kern-.05em{\sc i\kern-.025em b}\kern-.08em
    T\kern-.1667em\lower.7ex\hbox{E}\kern-.125emX}}
\begin{document}

\title{\title{ Recurrent Graph Convolutional Networks for Spatiotemporal Prediction of Snow Accumulation Using Airborne Radar}}

\author{\IEEEauthorblockN{Benjamin Zalatan}
\IEEEauthorblockA{\textit{Department of Computer Science and Engineering} \\
\textit{Lehigh University}\\
Bethlehem, PA, USA \\
bjz222@lehigh.edu}
\and
\IEEEauthorblockN{Maryam Rahnemoonfar}
\IEEEauthorblockA{\textit{Department of Computer Science and Engineering}\\
\textit{Department of Civil and Environmental Engineering}\\
\textit{Lehigh University}\\
Bethlehem, PA, USA \\
maryam@lehigh.edu}
}

\maketitle

\begin{abstract}
The accurate prediction and estimation of annual snow accumulation has grown in importance as we deal with the effects of climate change and the increase of global atmospheric temperatures. Airborne radar sensors, such as the Snow Radar, are able to measure accumulation rate patterns at a large-scale and monitor the effects of ongoing climate change on Greenland's precipitation and run-off. The Snow Radar's use of an ultra-wide bandwidth enables a fine vertical resolution that helps in capturing internal ice layers. Given the amount of snow accumulation in previous years using the radar data, in this paper, we propose a machine learning model based on recurrent graph convolutional networks to predict the snow accumulation in recent consecutive years at a certain location.  We found that the model performs better and with more consistency than equivalent nongeometric and nontemporal models.

\end{abstract}


\section{Introduction} 
As global atmospheric temperatures increase over time, the accurate prediction of annual snow accumulation and internal layers of ice sheets grows in importance. Tracking and forecasting these internal ice sheet layers is important for calculating snow mass balance, extrapolating ice age from direct measurements of the subsurface, and inferring otherwise difficult to observe ice dynamic processes. Precise understanding of the spatiotemporal variability of snow accumulation in the Greenland ice sheet is important to reducing the uncertainties in current climate model predictions and future sea level rise.

Measurements about the mass balance of ice sheets are traditionally collected by drilling ice cores and shallow pits. However, it is challenging to capture catchment-wide accumulation rates due to inherent sparsity, access difficulty, and high cost. Attempts to interpolate in-situ measurements further introduce uncertainties due to the high variability in local accumulation rate.

Airborne measurements collected over Greenland using nadir-looking radar sensors is one of the complementing methods of mapping topography and monitoring accumulation rates with a broad spatial coverage that has the advantage of revealing isochronous ice layers beneath the ice surface. Sub-surface ice layers that appear in snow radar echograms unveil historic annual and multi-annual snow accumulation. Snow Radar \cite{snow-radar} provides a spatiotemporal accumulation map that can help in quantifying the impact of atmospheric warming on polar ice caps and project its contribution to future sea level rise.

A large amount of airborne radar data from polar regions has been collected over the past few decades. However, the complexity, scale, and heterogeneity of the data has necessitated advanced automatic solutions. In recent years, several deep learning approaches \cite{rahnemoonfar2021deep,varshney2020deep,varshney2021deep,varshney2021refining, yari2021airborne, yari2019smart,yari2020multi,Yari-JSTARS,Varshney2022} based on convolutional neural networks (CNNs) have been proposed to track the internal layers of ice sheets and calculate historic snow accumulation. Snow Radar data, even after post-processing, contains a significant amount of noise. In addition to being very sensitive to noise, CNNs cannot be used for spatiotemporal tasks, such as predicting snow accumulation of the future years based on data from the past years.



Recent studies involving graph convolutional networks (GCNs) \cite{gcn} have shown promise in spatiotemporal tasks such as traffic forecasting \cite{relevant_traffic_rgcn} \cite{traffic1} \cite{traffic2}, wind speed forecasting \cite{wgatlstm}, and power outage prediction \cite{poweroutagegnn}. In this paper, we propose a geometric deep learning model that uses a supervised multi-target long short-term memory graph convolutional network (GCN-LSTM) \cite{gcnlstm} to predict the thickness of multiple shallow ice layers at specific coordinates given the thicknesses of multiple deep ice layers. 

\begin{figure}
    \centerline
    {
        \includegraphics[width=3cm, height=10cm]{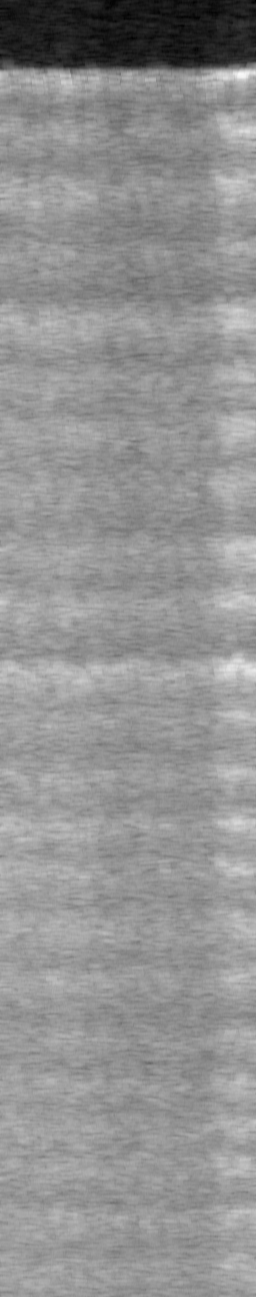}
        \hspace{0.75cm}
        \includegraphics[width=3cm, height=10cm]{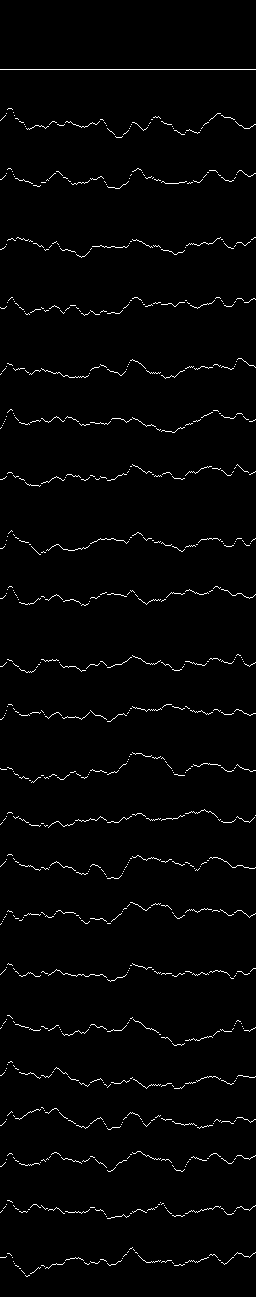}
        \includegraphics[height=10cm]{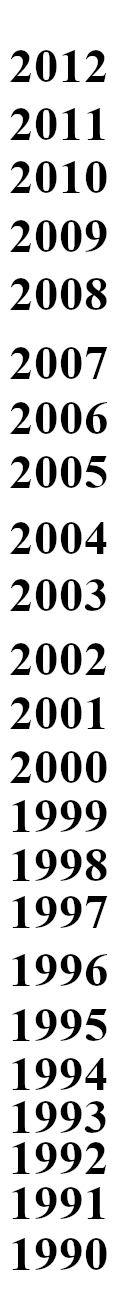}
    }
    \caption{Snow radar echogram (left) and its corresponding binary ground-truth labelling (right) showing layers of the Greenland ice sheet and their respective years.}
    \label{Fig1:radarAndGT}
\end{figure}

We focus on the Snow Radar \cite{snow-radar} dataset collected by the Center for Remote Sensing of Ice Sheets (CReSIS) as part of NASA's Operation IceBridge. The Snow Radar operates from 2-8 GHz and is able to track deep layers of ice with a high resolution over wide areas of an ice sheet. The sensor produces a two-dimensional grayscale profile of historic snow accumulation over consecutive years (see the left image in Figure \ref{Fig1:radarAndGT}), where the horizontal axis represents the along-track direction, and the vertical axis represents layer depth. Pixel brightness is directly proportional to the strength of the returning signal. Pixels representing surface layers are generally brighter and more well-defined due to high reflectance and snow density variation, while pixels representing deeper layers are generally darker and noisier due to increased density and a lower return-signal strength. In our experiments, we use radar data from selected Snow Radar flights over Greenland in the year 2012. In many areas, each ice layer represents an annual isochrone \cite{koenig2016annual}. As such, we may refer to specific ice layers by their corresponding year (see Figure \ref{Fig1:radarAndGT}).

In our experiments, we convert internal ice layer thickness information in radar images into sequences of temporal graphs to be used as input to our model. More specifically, we convert ten deep ice layers (corresponding to annual snowfall from 1997 to 2006, respectively) into ten spatiotemporal graphs. Our model then performs multi-target regression to predict the thicknesses of the five shallow ice layers immediately below the surface, corresponding to annual snow accumulation from 2007 to 2011, respectively.

Our contribution in this work includes the following:
\begin{enumerate}
    \item A method of converting Snow Radar ice layer data into sets of spatiotemporal graphs.
    \item A generalizable machine learning model based on GCN-LSTM that predicts the thicknesses of shallow ice layers.
\end{enumerate}

Our proposed model was shown to perform better with more consistency than equivalent non-geometric and non-temporal models.


\section{Related Work}

\subsection{Deep Learning for the Calculation of Ice Layer Thickness and Snow Accumulation}
Automated techniques, such as \cite{Rahnemoonfar-level-set, Rahnemoonfar-charged-particles, Rahnemoonfar_AIRadar}, have been developed for detecting the surface and bottom layers of echograms using radar depth sounder sensors.
Tracking the internal layers of an ice sheet is inherently more challenging due to the close proximity of each layer. Due to its exceptional performance in feature extraction, deep learning has been  applied extensively on remote sensing images, such as Snow Radar data, for tracking the internal layers of ice sheets \cite{rahnemoonfar2021deep} \cite{varshney2020deep} \cite{yari2019smart} \cite{yari2020multi}. \cite{yari2019smart} used a multi-scale contour-detection CNN to segment the different internal ice layers found in Snow Radar images. The authors of this work also experimented with pretraining on the Berkeley Segmentation Dataset and Benchmark (BSDS) dataset \cite{arbelaez2010contour} and found that this was not very effective due to the large amount of noise in Snow Radar images. In \cite{rahnemoonfar2021deep}, the authors trained a multi-scale neural network on synthetic Snow Radar images for more robust training. A multi-scale network was also used in \cite{yari2020multi}, where the authors trained a model on images from the year 2012 and fine tuned it by training on a small number of images from other years. \cite{varshney2020deep} found that using pyramid pooling modules, a type of multi-scale architecture, helps in learning the spatio-contextual distribution of pixels for a certain ice layer. The authors also found that denoising the images prior to the CNN layer improved both the accuracy and F-score. These works have all used multi-scale networks in order to segment the ice layers within Snow Radar images, but have commonly noted that the high amount of noise present in Snow Radar images is an issue that must be addressed in the future.

While these works have attempted to track and segment the annual ice layers present within Snow Radar images, none have yet attempted to use the spatiotemporal patterns of the ice layers to predict the thicknesses of shallow ice layers given the thicknesses of deeper ice layers.

\subsection{Graph Convolutional Networks}

Graph convolutional networks have had numerous real-world applications in a vast array of different fields. In the field of computer vision, recurrent GCNs have been used to generate and refine ``scene graphs'', where each node corresponds to the bounding box of an object in an image and the edges between nodes are weighted by a learned ``relatedness'' factor that prunes unrelated edges \cite{scenegraph1} \cite{scenegraph2}. GCNs have also been used to segment and classify point clouds generated from LiDAR scans \cite{lidar1} \cite{lidar2}. Recurrent GCNs have been used in traffic forecasting, such as in \cite{relevant_traffic_rgcn}, where graph nodes represented sensors installed on different roads, edges between nodes were weighted by the distance between sensors, and node features consisted of average detected traffic speed over a certain period of time.


Some existing recurrent graph-based weather prediction models, such as \cite{wgatlstm} and \cite{wgcnlstm}, have experimented with defining the edge weights in graph adjacency matrices as learnable parameters rather than static values. While this did show slightly improved performance by allowing a model to learn relationships between nodes more complex than simple geographic distance, it exponentially increases the computational complexity of the model, making it considerably more difficult to train and increasing its susceptibility to overfitting. While it is possible to reduce the number of learnable parameters if the adjacency matrix is low-rank, this is not always guaranteed and the number of learnable parameters may remain very high. This number may also be decreased by using a more sparse graph structure, such as restricting a node's neighborhood to its $k$-nearest neighbors. This strategy can be applied for some problems, but may not work for spatial tasks where most or all nodes are in close proximity.

\section{Dataset}
\begin{table}
    \caption{Key parameters of the Snow Radar used during data collection.} 
    \begin{centering}
    \label{table:SnowRadarParam}
    \begin{tabular}{lll}
    $\quad$ &$\quad$& Snow Radar Parameters \\
    \hline
    Bandwidth  &$\quad$& 2-8 GHz \\
    Pulse duration & $\quad$ & 250 $\mu$s \\
    PRF &$\quad$& 2 kHz\\
    Transmit power  &$\quad$& 100 mW \\
    Intermediate frequency range  &$\quad$& 62.5-125 MHz \\
    Sampling frequency  &$\quad$&  125 MHz \\
    Range resolution  &$\quad$& $ \sim{4} $ cm \\
    Along-track footprint  &$\quad$& 14.5 m
    \end{tabular}
    \end{centering}
    \end{table}
In this study, we used Snow Radar data captured during the NASA's Operation Ice Bridge mission \cite{snow-radar}. CReSIS has made this data publicly available on their website (\url{https://data.cresis.ku.edu/}). The data gathered during each flight was processed and separated into a series of grayscale radar echogram images, each with a width of $256$ pixels and a height ranging between $1200$ and $1700$ pixels. Each pixel in a column corresponds to approximately $4$ cm of ice, and each radar image has an along-track footprint of 14.5m (see Table \ref{table:SnowRadarParam}). The value of each pixel is proportional to the relative received power of the returning radar signal at that depth. Accompanying each image was a $256 \times 2$ matrix specifying the geographic latitude and longitude associated with each vertical column of pixels. In order to gather ground-truth thickness data, the images were manually labelled in a binary format where white pixels represented the tops of each firn layer, and all other pixels were black (see the right image in Figure \ref{Fig1:radarAndGT}).

In order to capture a sufficient amount of data, only radar images containing a minimum of $16$ ice layers were used: one unused surface layer, five shallow layers for ground truths, and ten deeper layers for use as node features (the surface sheet is unused as it remains flat with constant height throughout all images in the dataset). Ten deep layers and five shallow layers were chosen in order to maximize the number of usable images while also maintaining a reasonable amount of data per image. This limitation reduced the total number of usable images from $816$ down to $568$. Five different training and testing sets were generated by taking five random permutations of all usable images and splitting them at a ratio of 4:1. The five training sets contained $454$ images each, and the five testing sets contained $114$ images each.

\begin{figure*}
    \centerline
    {
        \includegraphics[width=\textwidth]{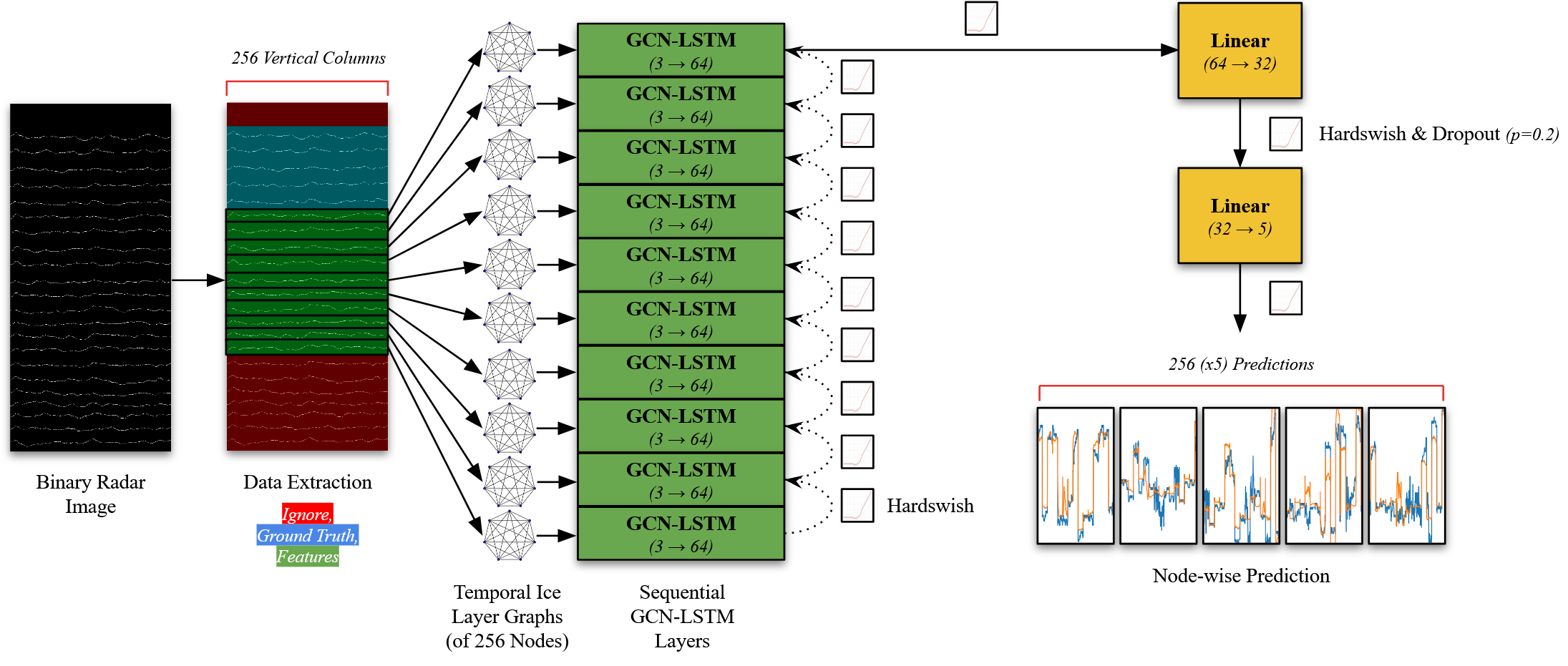}
    }
    \caption{Architecture of the proposed model.}
    \label{Fig2:arch}
\end{figure*}

\section{Methods}


\subsection{Graph Convolutional Networks}

Traditional CNNs use a matrix of learnable weights, often referred to as a kernel or filter, as a sliding window across pixels in an input image. This allows a model to automatically extract image features that would otherwise need to be identified and inputted manually. Graph convolutional networks apply similar logic to graphs, but rather than using a sliding window of weights across an image, GCN performs weighted-average convolution on each node's neighborhood to extract features that reflect the structure of a graph. The size of the neighborhood on which convolution takes place is dictated by the number of GCN layers present in the model (i.e. $K$ GCN layers result in $K$-hop convolution). In a sense, graph convolutional networks are a generalized form of traditional CNN's that enable variable degree and weighted adjacency.

\subsection{Recurrent Neural Networks}

Recurrent neural networks (RNNs) are able to process a sequence of data points as input, rather than a single static data point, and learn the long-term relationships between them. Many traditional RNN structures had issues with vanishing and exploding gradients on long input sequences. Long short-term memory (LSTM) attempts to mitigate those issues by implementing gated memory cells that guarantee constant error flow \cite{lstmpaper}. Applying LSTM to GCN using GCN-LSTM allows for a model to learn not only the relationships between nodes in a graph, but also how those relationships change (or persist) over time.

\subsection{Model Architecture}

Our model uses ten consecutive GCN-LSTM layers (one for each of the ten deep ice layers from 1997 to 2006), each with $64$ output channels, that lead into two fully-connected layers. The first fully-connected layer has $32$ output channels, and the second fully-connected layer has five output channels. The output of the model is the predicted thickness (in pixels) of the five shallow ice layers (2007 to 2011) immediately beneath the surface. Between each layer is the Hardswish activation function \cite{hardswishpaper}, chosen due to its superior performance when compared to ReLU and its derivatives \cite{swishbetter}. Between the two fully-connected layers is Dropout \cite{dropoutpaper} with $p=0.2$. We use the Adam optimizer \cite{adampaper} with a constant learning rate of $0.01$, and we train for $150$ epochs using mean-squared error loss. The structure of the model is shown visually in Figure \ref{Fig2:arch}.

\subsection{Graph Generation}


\subsubsection{Model Input}

Each radar image is converted into ten graphs, each consisting of $256$ nodes. Each graph corresponds to a single ice layer for each year from 1997 to 2006. Each node represents a vertical column of pixels in the radar echogram image and has three features: two for the latitude and longitude that correspond to that column, and one for the thickness of the corresponding year's ice layer within that column.

\subsubsection{Graph Adjacency}

All graphs are fully connected and undirected. All edges are inversely weighted by the geographic distance between node locations using the haversine formula. For a weighted adjacency matrix $A$:

{\small
    \begin{equation*}
        A_{i, j} = \frac{1}{2\arcsin\bigg(\text{hav}(\phi_j - \phi_i) + \cos(\phi_i)\cos(\phi_j)\text{ hav}(\lambda_j - \lambda_i)\bigg)}
    \end{equation*}
}
where
{\small
\begin{equation*}
    \text{hav}(\theta) = \sin^2 \bigg(\frac{\theta}{2} \bigg)
\end{equation*}
}
$A_{i, j}$ represents the weight of the edge between nodes $i$ and $j$. $\phi$ and $\lambda$ represent the latitude and longitude features of a node, respectively. Node features of all graphs are collectively normalized using z-score normalization. Weights in the adjacency matrices of all graphs are collectively normalized using min-max normalization.

\section{Results}

\begin{table*}
    \centering
    \label{table:OverallResults}
    \caption{Results from the LSTM, GCN, and GCN-LSTM models on the five predicted annual ice layer thicknesses from 2007 to 2011. Results are shown as the mean $\pm$ standard deviation of the RMSE over five trials (in pixels).}
    \begin{tabular}{ | c | c | c | c | c | c | c | } 
        \hline
        & 2007 & 2008 & 2009 & 2010 & 2011 & Total \\ 
        \hline
        LSTM & $7.834 \pm 1.050$ & $8.620 \pm 3.392$ & $7.243 \pm 1.213$ & $5.653 \pm 0.370$ & $5.995 \pm 0.764$ &  $7.268 \pm 1.162$ \\ 
        \hline
        GCN & $6.165 \pm 0.431$ & $6.458 \pm 0.866$ & $6.376 \pm 1.381$ & $5.267 \pm 0.184$ & $5.184 \pm 0.768$ &  $5.956 \pm 0.468$ \\ 
        \hline
        GCN-LSTM & $\mathbf{4.614 \pm 0.504}$ & $\mathbf{4.733 \pm 1.111}$ & $\mathbf{5.278 \pm 1.074}$ & $\mathbf{4.449 \pm 0.336}$ & $\mathbf{3.855 \pm 0.272}$ & $\mathbf{4.651 \pm 0.414}$ \\ 
        \hline
    \end{tabular}
\end{table*}

\subsection{Baselines}

To highlight the performance of the GCN-LSTM model,  as well as verify that the temporal and geometric aspects of the model serve to its benefit, we compared its performance with equivalent models that utilize a traditional GCN as well as a traditional, non-geometric LSTM.

For the GCN model, all hyperparameters remain the same, but the GCN-LSTM layer sequence is replaced by a single GCN layer. Rather than generating ten graphs for each of the ten deep ice layers, we generate a single graph with each node having 12 features: two for the latitude and longitude, and ten for the thicknesses of the 1997 to 2006 deep ice layers. The rest of the model, including the adjacency matrix generation, is identical to the proposed model.

\begin{figure}[H]
    \centering
    \begin{tabular}{l l}
    \includegraphics[width=3.5cm, height=4cm]{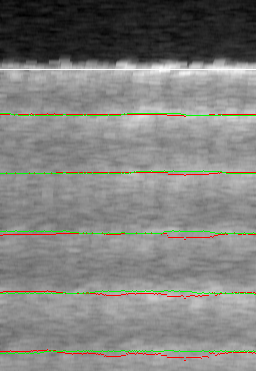}
    \includegraphics[width=3.5cm, height=4cm]{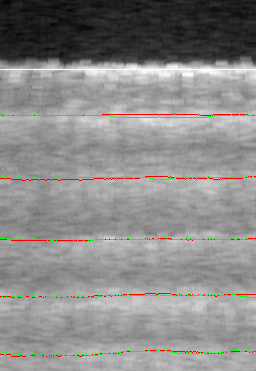}
    \\
    \includegraphics[width=3.5cm, height=4cm]{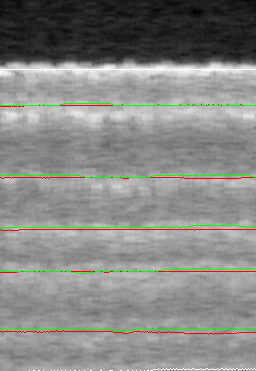}
    \includegraphics[width=3.5cm, height=4cm]{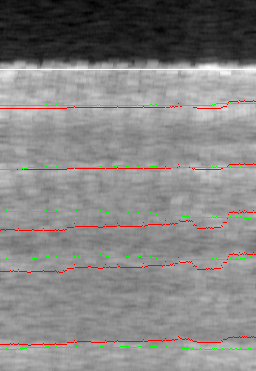}
    \end{tabular}
     \caption{Example radar images overlayed with predicted thicknesses from our model. Green pixels are the ground truth, red pixels are the predicted value. A poor-fitting example is shown in the bottom-right.}
     \label{Fig3:overlays}
\end{figure}

For the LSTM model, all hyperparameters remain the same, but ten sequential LSTM layers are used, rather than ten sequential GCN-LSTM layers. Since this model is non-geometric, nodes are simply converted into rows in a feature vector, and no adjacency information is supplied. The rest of the model is identical to the proposed model.

\begin{figure*}[ht!]
    \centering
    \begin{tabular}{l l}
    \begin{tabular}{@{}c@{}}
    \includegraphics[width=8.5cm]{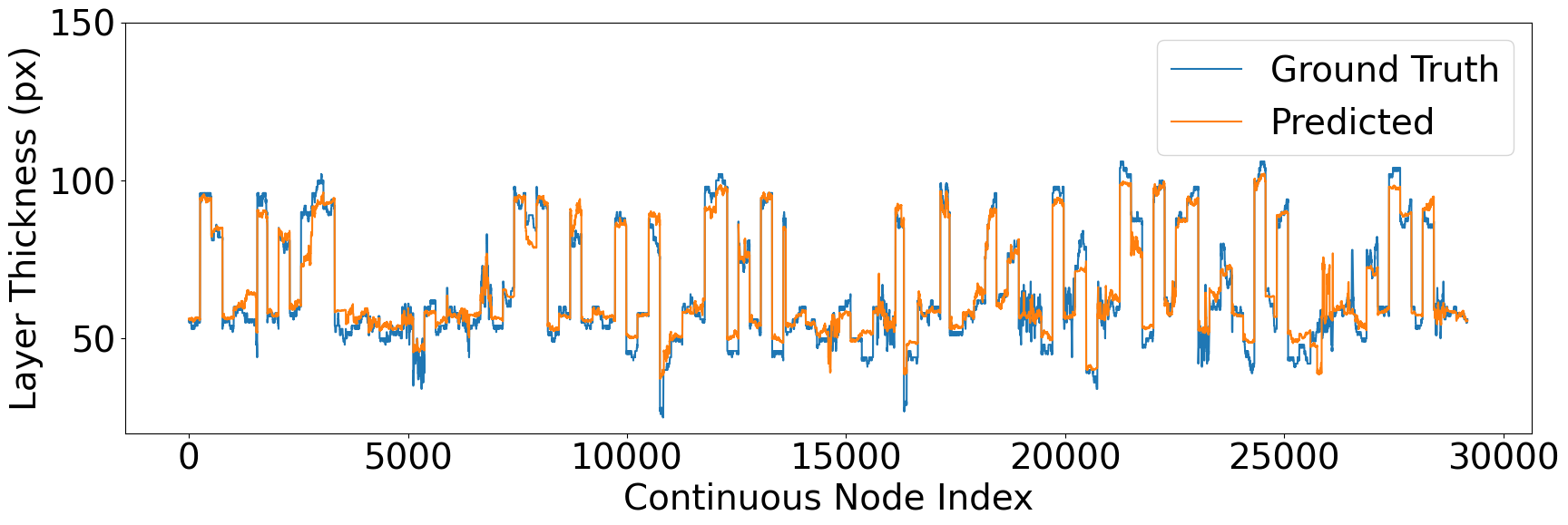} \\ \footnotesize{(a) Continuous predicted thicknesses for 2007.}
    \end{tabular} &
    \begin{tabular}{@{}c@{}}
    \includegraphics[width=8.5cm]{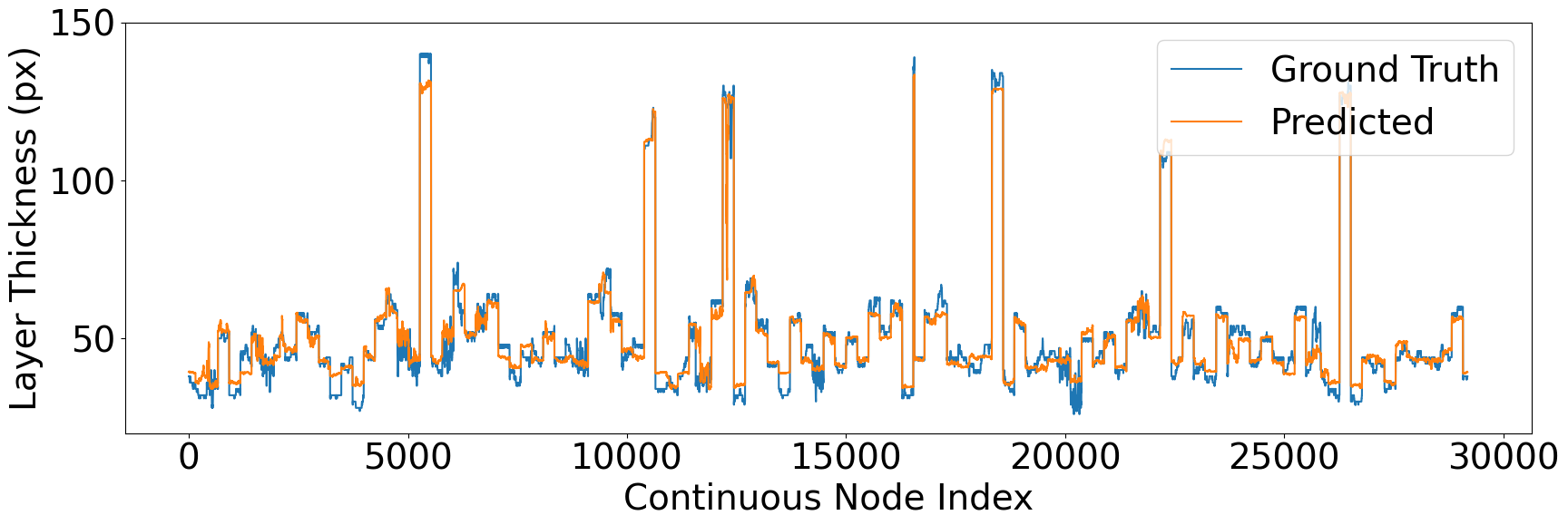} \\ \footnotesize{(b) Continuous predicted thicknesses for 2008.}
    \end{tabular} \vspace{0.25cm}\\
    \begin{tabular}{@{}c@{}}
    \includegraphics[width=8.5cm]{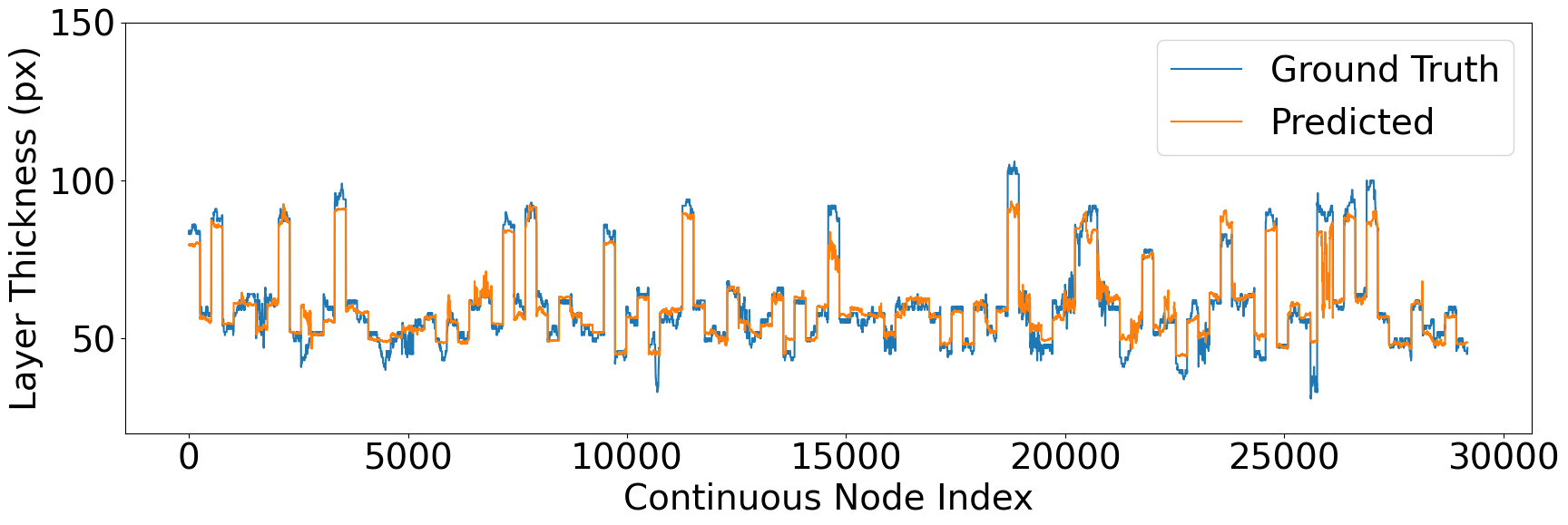} \\ \footnotesize{(c) Continuous predicted thicknesses for 2009.}
    \end{tabular} &
    \begin{tabular}{@{}c@{}}
    \includegraphics[width=8.5cm]{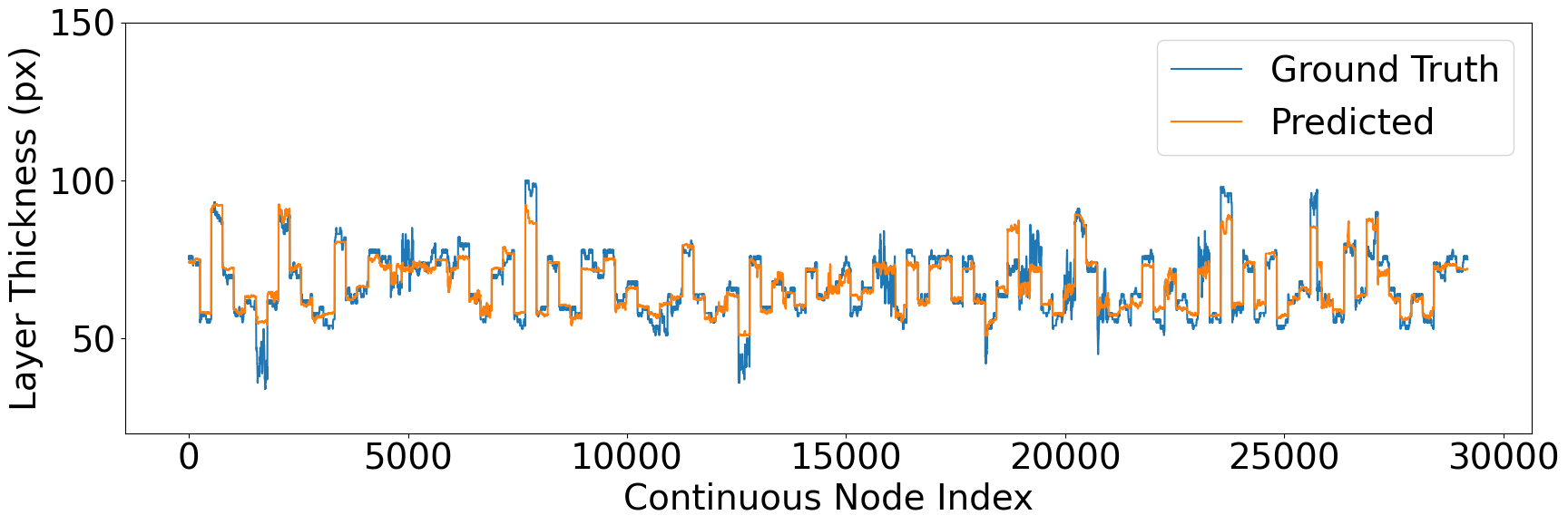} \\ \footnotesize{(d) Continuous predicted thicknesses for 2010.}
    \end{tabular} \vspace{0.25cm}\\
    \multicolumn{2}{c}{
    \begin{tabular}{@{}c@{}}
    \includegraphics[width=8.5cm]{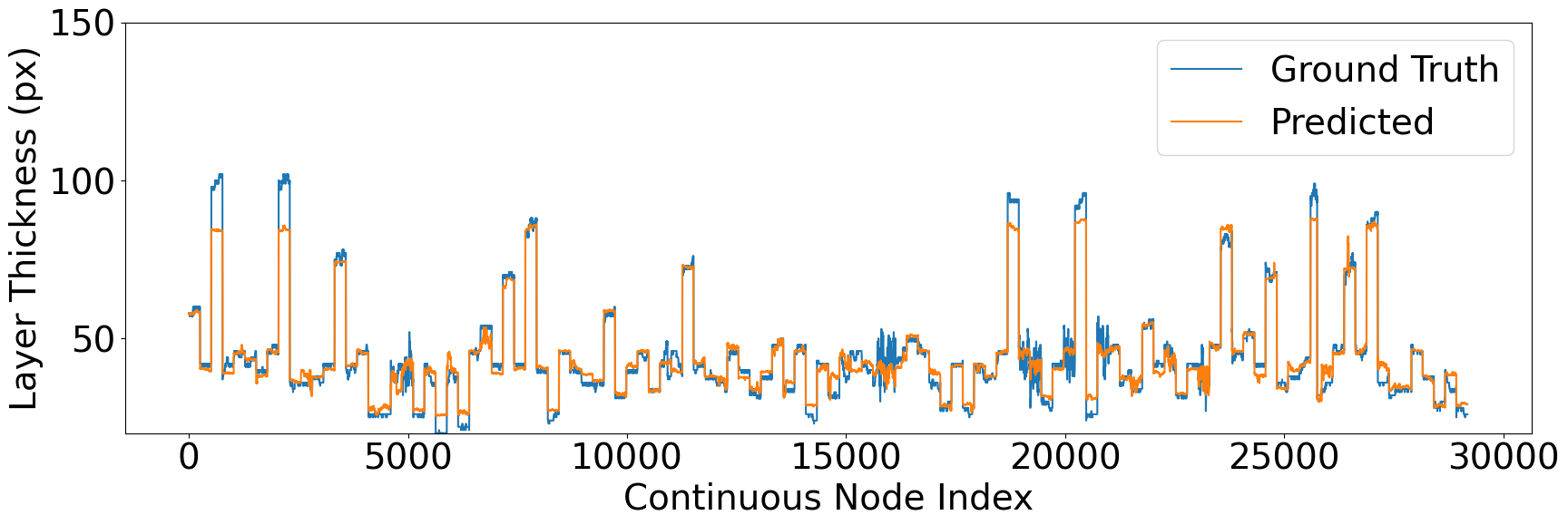} \\ \footnotesize{(e) Continuous predicted thicknesses for 2011.}
    \end{tabular}
    }\\
    \end{tabular}
    \caption{Continuous real vs. predicted thicknesses for years 2007-2011. Ground truths are labelled in blue, predicted values are labelled in orange. The $y$-axis represents ice layer thickness in pixels, and the $x$-axis represents the unsorted, randomized node index in the testing set.}
    \label{Fig4:continuous}
\end{figure*}

\subsection{Experimental Results}

In each trial, the root mean squared error (RMSE) was taken between the predicted and ground truth thicknesses for each of the 2007-2011 ice layers among the entire testing set. The mean RMSE and standard deviations were found for each year over all five trials. These results are showcased in Table II. Example well-fitting and poor-fitting results are overlayed on their respective radar echogram images in Figure \ref{Fig3:overlays}. Graphed continuous results over all (unordered) testing samples for years 2007-2011 are shown in Figure \ref{Fig4:continuous}. The proposed GCN-LSTM model performed significantly better than the baseline GCN and LSTM models both in terms of mean RMSE and consistency (shown by a lower standard deviation). The lack of adjacency data in the LSTM model and complex temporal learning in the GCN model may contribute to these results.

        
        

\section{Conclusion}



In this work, we proposed a temporal, geometric, multi-target machine learning model based on GCN-LSTM that predicts the annual snow accumulation of Greenland from 2007 to 2011 given the annual snow accumulation from 1997 to 2006. Our proposed model was shown to perform better and with more consistency than equivalent non-geometric and non-temporal models. While our model succeeds at predicting shallow layer thicknesses with reasonable accuracy, there are opportunities for improvement and generalization, some of which are outlined in the next section.

\subsection{Generalizability and Future Work}

\subsubsection{Additional Ice Layers}

While our model was restricted to learning on only ten ice layers and predicting only five ice layers, it is easily expandable to a higher number of both feature and predicted layers. Future work may take advantage of this expandability and test a similar model on a much larger number of layers.

\subsubsection{Additional Node Features}

In our model, minimal node features are present; only latitude, longitude, and layer thickness are used as node features. It may be possible to improve performance by integrating additional node features, such as physical properties of the ice at each layer. 

\subsubsection{Learnable Graph Adjacency}

In our model, edge weights between nodes are static values inversely proportional to their geographic distance. It may be possible to improve performance by defining edge weights as learnable parameters, allowing the model to learn relationships between nodes more complex than simple geographic distance.

\subsubsection{Generalizing to Future Snow Accumulation}

In these experiments, our model is restricted to learning and predicting on data gathered about existing ice layers. If radar image data gathered over multiple years were used, our model could be generalized to forecast future snow accumulation.

\subsubsection{Generalizing to Deep Ice Layers}

Since Snow Radar echogram images inherently become noisier and more difficult to track as depth and firn density increases, it may be useful to attempt to reverse this model and use the thicknesses of shallow ice layers to predict the thicknesses of deeper ice layers.

\section{Acknowledgements}
This work is supported by NSF BIGDATA awards (IIS-1838230, IIS-1838024), IBM, and Amazon. We acknowledge data and/or data products from CReSIS generated with support from the University of Kansas and NASA Operation IceBridge.

\bibliographystyle{unsrt}
\bibliography{references}

\end{document}